\documentclass[10pt]{article}
\usepackage{multicol}
\usepackage{esvect}
\usepackage{amsmath}
\usepackage{bbm}
\usepackage{amssymb}
\usepackage{graphicx}
\usepackage{caption}
\usepackage{tabularx}
\usepackage{float}
\usepackage{subcaption}
\usepackage{multirow}
\usepackage{bbm}
\usepackage{url}
\usepackage{paralist}
\usepackage[letterpaper, margin=0.75in]{geometry}

\title{Using Context to Improve Word Segmentation}
\author{Stephanie Hu, Xiaolu Guo}
\date{}

\begin{document}

\maketitle

\begin{multicols*}{2}
\section{Abstract}
An important step in understanding how children acquire languages is studying how infants learn word segmentation. It has been established in previous research that infants may use statistical regularities in speech to learn word segmentation. The research of Goldwater \textit{et al.}, demonstrated that incorporating context in models improves their ability to learn word segmentation \cite{goldwater}. We implemented two of their models, a unigram and bigram model, to examine how context can improve statistical word segmentation. The results are consistent with our hypothesis that the bigram model outperforms the unigram model at predicting word segmentation. Extending the work of Goldwater \textit{et al.}, we also explored basic ways to model how young children might use previously learned words to segment new utterances. 

\section{Introduction}
As infants acquire language, they must first understand the nature of words. Because most speech consists of words spoken consecutively without pauses in between, this raises the interesting question of how children learn to identify the boundaries between words. Many researchers have sought to understand how children learn word segmentation. Evidence from various research indicate that infants may use statistical regularities in strings of syllables as cues for word segmentation \cite{saffran}. The observation that it is harder to predict the next syllable between two words than within a word has led many researchers to disregard the importance of context in learning word segmentation \cite{harris, saffran}. However, Goldwater \textit{et al.} suggest that context may actually play a role in the process of learning word segmentation \cite{goldwater}.

Building on top of previous research in statistical word segmentation, Goldwater \textit{et al.} proposed the importance of understanding the assumptions the child makes about speech, and in particular, the nature of a word. Specifically, they suggested two different assumptions about the nature of a word: 1) a word occurs statistically independent of other words, and 2) a word can be used to predict other words \cite{goldwater}. 

To model the problem, we can view each word as a composition of syllables, or phonemes. We seek to better understand how children find boundaries in a sequence of phonemes and turn such a sequence into segmented words. For this paper, we investigate the impact of using context, e.g. the surrounding words, on learning word boundaries. 

We define an utterance as a sequence of words $w_1, w_2,..,w_k$ followed by a pause after the $k$th word. Based on the assumption, used by Goldwater \textit{et al.} that an utterance can be modeled as a Dirichlet process, we created two types of models that learn word segmentation: a unigram model and a bigram model. In the paper, we will also discuss how our models can also be derived from the Chinese restaurant process (CRP)  \cite{goldwater}. Although these models are ideal learners (in other words, they will yield the best possible results given the input data), they can be used to explore the impact of using previous words in predicting word boundaries. 

A unigram is single word while a bigram is a pair of words that have a sequential relationship. The unigram model reflects the belief that context is irrelevant to learning word segmentation and that each word is generated independent of other words. The bigram model encapsulates the belief that each word is generated dependent on the previous word. Previous research by Goldwater \textit{et al.} shows that the bigram model outperforms the unigram model in the word segmentation task, indicating that context is indeed relevant \cite{goldwater}. 

In the rest of this paper we will discuss the unigram and bigram models that are proposed for generative utterances. We  also analyze our inference processes and the results of learning word segmentation with our unigram and bigram models. Finally, we  discuss challenges we faced, improvements to our models, and possible extensions to our research. 

\vspace{-1em}
\section{Models}
The research of Goldwater \textit{et al.} suggests that context helps infants learn word segmentation. In order to further investigate this hypothesis, we examined two models of learning word segmentation: a unigram model and a bigram model. We first describe these models using two different views: the Chinese restaurant process (CRP) and the Dirichlet process (DP). Then we provide explicit details for how word and utterance generation occur under each model. We also briefly discuss two previous related models, NGS-u ($n$-gram Segmentation System, unigram) and MBDP-1 (Model Based Dynamic Programming) and how they compare to the context model \cite{brent, ngs}. 

\subsection{Unigram model}

The unigram model assumes that each word is generated independently, and thus does not incorporate dependencies on surrounding words. We first examine this model in order to compare it to other accepted models like NGS and MBDP-1 that lack the notion of context.

\subsubsection{Chinese Restaurant Process}
CRP is used to partition units into separate groups \cite{crp}. Suppose there is a restaurant with infinite tables, where each table has infinite seats. Each customer will sit at an empty table with some probability proportional to some parameter $\alpha_0$. Otherwise, the customer will sit at an already occupied table with some probability proportional to the number of people already sitting at that table. The parameter $\alpha_0$ then influences how many tables are occupied. In the context of generating sequences of words, each table represents a single word, with the words generated by some base distribution, $P_0.$ The customers at each table represents the number of occurrences of that word in the sequence so far \cite{goldwater}.

\subsubsection{Dirichlet Process}
Goldwater \textit{et al.} proposed that an utterance $u$ consisting of a sequence of words $w_1 w_2 ...w_k$ can be also modeled as a Dirichlet process. Dirichlet distributions are often used to introduce the concept of a prior in Bayesian inference, and the outcome of sampling from this distribution is another distribution \cite{goldwater}. The Dirichlet process is a nonparametric model characterized by a concentration parameter $\alpha_0$ that establishes the sparsity of the resulting distribution and a base distribution $P_0$ that determines the expected value of the process. In the context of generating utterances, the value of $\alpha_0$ determines the expected number of unique word types in a given sequence of word tokens while $P_0$ provides insight into the particular phonemes that form a specific word type.  

\subsubsection{Generative Model for Words} 
Our unigram model assumes that an utterance is generated via a Dirichlet process (DP) \cite{goldwater}. Specifically, the $i$th word in the sequence $w_i$ is formulated as follows:
\setdefaultleftmargin{10pt}{}{}{}{}{}
\begin{enumerate}
\item Decide if $w_i$ is a new (previously unseen) word.
\item Depending on whether $w_i$ is a new word or not, perform the following
\begin{enumerate}
\item If $w_i$ is a new word, generate a sequence of phonemes ($x_1...x_m$) for $w_i.$
\item If $w_i$ is not a new word, choose an existing word $l$ in the lexicon for $w_i$.
\end{enumerate}
\end{enumerate}
We can then compute the probabilities for each decision in this generative model.
\setdefaultleftmargin{10pt}{}{}{}{}{}
\begin{enumerate}
\item P($w_i$ is new) = $\frac{\alpha_0}{n+\alpha_0}$, P($w_i$ is not new) = $\frac{n}{n+\alpha_0}$
\item 
\begin{enumerate}
\item $P(w_i=x_1...x_M|w_i \text{ is new})$ = 

\hspace{3em} $p_\#(1-p_\#)^{M-1}\prod^{M}_{j=1}P(x_j)$
\item $P(w_i=l | w_i \text{ is not new}) = \frac{n_l}{n}$
\end{enumerate}
\end{enumerate}
We define $n$ as the total number of words generated before $w_i$ (so that $n = i - 1$) and $n_l$ as the total number of times word type $l$ has occurred in those $n$ words. We also introduce $p_\#$, the probability of generating a phoneme that terminates a word (a word boundary). Under this model, we assume that each phoneme is generated independently at random from a uniform distribution, so the probability of generating a word comprised of $m$ phonemes is precisely $p_\#(1 - p_\#)^{m - 1}$. Note that $\alpha_0$ is the concentration parameter defined previously. Given these variable definitions, the probability $\mathbf{w}_{-i} = {w_1,...,w_{i-1}} $ of generating $w_i = l$ given the previous words in the sequence is
\[P(w_i = l|\mathbf{w}_{-i} = \frac{n_l}{i-1+\alpha_0}+\frac{\alpha_0P_0(w_i = l)}{i-1+\alpha_0}\] where $P_0$ is the base distribution over generating a unigram sequence of phonemes ($w_i= x_1,...x_M)$ that was defined in step 2a \cite{goldwater}. 

\subsubsection{Utterance Boundaries} 
Now that we have a model for producing a sequence of words, we also need to consider the process of generating the end of that sequence, e.g. an utterance boundary. To model utterance we use the following generative procedure: 
\begin{enumerate}
\item Decide if the next word $w_i$ will end the utterance with probability $p_\$$. In other words, decide if $p_\$$ is utterance-final or not. 
\item Choose the word type $w_i = l$ according to the DP model discussed previously. 
\item If $w_i$ is utterance-final, terminate the utterance. Otherwise, return to the first step. 
\end{enumerate}

\noindent  
The parameter $p_\$$ is unknown, so in accordance with the Goldwater paper, we assume that the value of $p_\$$ is drawn from a symmetric Beta($\frac{\rho}{2}$) prior \cite{goldwater}. Then the probability that $w_i$ is utterance-final is given by 
$$P(u_i = 1 | \mathbf{u}_{-i}, \rho) = \frac{n_\$ + \frac{\rho}{2}}{i - 1 + \rho}$$

\noindent  
Here, $u_i$ is a binary random variable where $u_i = 1$ if $w_i$ is utterance-final and $u_i = 0$ otherwise. In addition, $\mathbf{u}_{-i}$ is the sequence of $u_j$ values from $j = 0...i - 1$. The derivation of this formula is given in the Goldwater paper, so we exclude the computation here \cite{goldwater}.  

\subsection{Bigram model}
We will first extend the CRP and DP interpretations of utterance generation to fit the bigram model.

\subsubsection{Hierarchical Chinese Restaurant Process}
We can use a hierarchical CRP to model bigrams \cite{goldwater}. Instead of just one restaurant, we now have an infinite number of restaurants and one ``master'', or ``backoff'', restaurant. The backoff restaurant is similar to the restaurant we described in the CRP view of the unigram model: each table in the backoff restaurant is labeled with some word drawn from a base distribution $P_0$. Each customer sitting at a table in the backoff restaurant represents the occurrence of the label of that table in a bigram. Next, there is a restaurant for each word $l$, where $l$ represents the first word in a bigram. Each of the tables $t$ in the restaurant represents a word generated from the distribution in the backoff restaurant; this is the second word in the bigram. Finally, each customer sitting at a table in the restaurant $l$ represents the number of occurrences of the bigram $\langle l, t \rangle$.

\subsubsection{Hierarchical Dirichlet Process}
We will consider a hierarchical DP to fit our bigram model \cite{goldwater}. In the unigram DP model, we introduced parameters $\alpha_0$ and $P_0$ (see Section 3.1.2 for details). For our hierarchical DP, we add a new distribution $G$ that is generated from another DP. This distribution $G$ serves as a base distribution that links together the DP $H_l$ for a specific lexicon $l$ to all other words. Note that $P(w_i|w_{i-1} = l)$, or the probability of $w_i$ conditionally dependent on the previous word $w_{i - 1}$, is distributed according to $H_l$ \cite{goldwater}. Finally, we introduce an additional parameter of the model $\alpha_1$ that determines the probability of generating a novel bigram. 

\subsubsection{Generative Model}
For the bigram model, we assume the following generative model given in the Goldwater paper \cite{goldwater}: in any sequence of words, given the previous word $w_{i-1}$, the $i$th word $w_i$ is generated as follows: 
\begin{enumerate}
\item Decide if $\langle w_{i-1},w_i\rangle$ is a new bigram type.
\item 
\begin{enumerate}
\item If $\langle w_{i-1},w_i\rangle$ is a new bigram, \begin{enumerate}
\item Decide if $w_i$ is a new unigram type.
\item \begin{enumerate}
\item If $w_i$ is a new word, generate a sequence of phonemes ($x_1...x_m$) for $w_i.$
\item If $w_i$ is not a new word, choose an existing word for $w_i.$
\end{enumerate}
\end{enumerate}
\item If $\langle w_{i-1},w_i \rangle$ is not a new bigram, choose an existing word for $w_i$ from those that have been previously generated after generating $w_{i-1}.$
\end{enumerate}
\end{enumerate}
We can then formulate the corresponding probabilities for each decision in the generative model. 
\begin{enumerate}
\item $P(\langle w_{i-1},w_i\rangle \text{ is new}|w_{i-1} = l') = \frac{\alpha_1}{n_{l'}+\alpha_1}$

\hspace{0.2em}$P(\langle w_{i-1},w_i\rangle \text{ is not new}|w_{i-1} = l') = \frac{n_{l'}}{n_{l'}+\alpha_1}$
\item 
\begin{enumerate}
\item 
\begin{enumerate}
\item $P(w_i\text{ is new}|\langle w_{i-1},w_i\rangle \text{ is new}) = \frac{\alpha_0}{b+\alpha_0}$

\hspace{0.2em}$P(w_i\text{ not new}|\langle w_{i-1},w_i\rangle \text{ is new}) = \frac{b}{b+\alpha_0}$
\item 
\begin{enumerate}
\item $P(w_i=x_1...x_M|w_i \text{ is new}) = $

$P_0(x_1,...,x_M)$

\item $P(w_i= l|w_i \text{ is not new}) = \frac{b_l}{b}$  
\end{enumerate}
\end{enumerate}
\item $P(w_i=l|\langle w_{i-1},w_i\rangle \text{ not new}, w_{i-1} = l') = \frac{n_{\langle l',l\rangle}}{n_{l'}}$
\end{enumerate}
\end{enumerate}
Here, $P_0$ refers to the same base distribution as in the unigram model. $\alpha_0$ and $\alpha_1$ are parameters of the model. We define $l'$ to be the sequence of phonemes that makes up the $i$th word $w_i$. We define $n_l$ and $n_{\langle l',l\rangle}$ as the number of times the lexicon $l$ and the number of times the bigram $\langle l',l\rangle$, respectively, are generated in the words before $w_i$ in the sequence. Finally, $b$ is the number of bigram types, whiel $b_l$ is the number of bigram types whose second word is $l$ in the first $i-1$ words. 

\subsection{Other Models} 
In this section we briefly discuss two other well-known models for word segmentation: the NGS models proposed by Venkataraman (2001) and the MBDP-1 model proposed by Brent (1999). 

Venkataraman implemented three NGS models using unigrams, bigrams, and trigrams, respectively. These models and the MBDP-1 model proposed by Brent use online search procedures and produce relatively noisy results, but their models appear to stabilize quickly \cite{brent, ngs}. It should be noted that the optimal solution under the NGS models is a completely undersegmented corpus, and the segmentation accuracies produced by all three of Venkataraman's models are similar. This suggests that contextual dependencies are not important for proper word segmentation. The performances of both the NGS and MBDP-1 models are influenced by the search algorithm used \cite{goldwater}. 

\section{Inference} 

We aimed to solve the following inference problem in our project: given a sequence of phonemes, separate the sequence into distinct words. We use Gibbs sampling to learn the posterior distribution of lexicons and simultaneously determine the word boundaries in the input sequence of phonemes. (Recall that this posterior distribution is formulated using the model presented earlier.) The inference procedure described here is based off the work of Goldwater \textit{et al.} In their models and ours, we assume that the utterance boundaries are given. This is a logical assumption since there is generally some kind of pause in speech that can serve as a signal for a boundary point. Thus, the segmentation problem is reduced to learning the correct boundaries for words within an utterance.

\begin{figure*}[!b]
    \centering
    \includegraphics[width=0.8\textwidth]{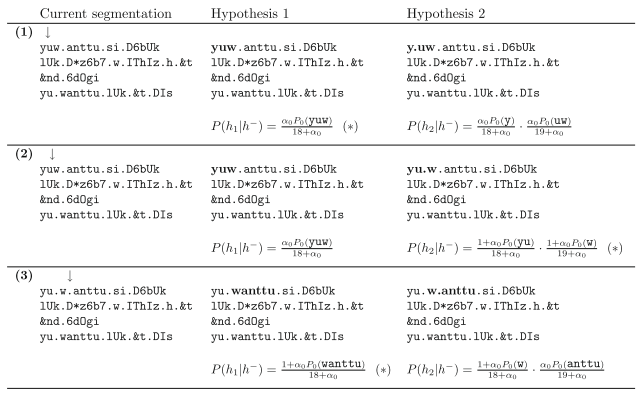}
    \caption{Sample iterations of the Gibbs sampling algorithm under the unigram model. Figure originally appeared in Goldwater \textit{et al.}, 2009.}
    \label{fig:sample-gibbs}
\end{figure*} 

\subsection{Gibbs Sampling}
Gibbs Sampling is a Markov chain Monte Carlo algorithm in which we sample each variable from their conditional posterior distribution in turn, while holding the values of all other variables in the model constant at their current values. In our experiment with word segmentation, we consider a single potential word boundary location while fixing the locations of all other current boundaries in the corpus. This amounts to sampling from the posterior distribution $P(h|d)$, where $d$ is the data (unsegmented corpus) and $h$ is a hypothesis in our hypothesis space. 

In fact, we only have two possible hypotheses to consider. We refer to $h_1$ as the hypothesis that there is no boundary at the current index $i$, e.g. there is no word that ends at index $i$. On the other hand, $h_2$ is the hypothesis that there is a boundary at index $i$, thus splitting the sequence of phonemes at $i$ into a word on the left and a word on the right. At the start of the sampling procedure, we initialize the boundaries at random since Goldwater \textit{et. al.} found that initialization did not affect the final results. This makes sense because sampled values will eventually converge to the true posterior distribution after many successive iterations. Note that in our implementation, we discard the first thousand iterations through the training data to allow the sampler to start converging to the true posterior distribution before collecting samples for the actual inference. This is known as the ``burn-in'' period. 

\subsection{Inference Under Unigram Model}
During each round of our Gibbs sampling procedure, we consider each possible boundary location in the input sequence of phonemes and compute $P(h_1 | d)$ and $P(h_2 | d)$. We then select one of the hypotheses with probability proportional to their relative conditional probabilities. 

Consider the potential boundary location $i$. We could choose to not insert a boundary, in which case our segmented corpus would look like $\beta w_1 \delta$, where $w_1$ is a word that spans across index $i$ and $\beta$ and $\delta$ represents the segmented phoneme sequences to the left and right of $w_1$, respectively. This corresponds to choosing $h_1$. On the other hand, if we choose to insert a boundary (e.g. if we choose $h_2$), then our corpus becomes $\beta w_2 w_3 \delta$, where index $i$ falls between $w_2$ and $w_3$ and $w_1 = w_2 w_3$. We can compute $P(h_1 | h^-, d)$ and $P(h_2 | h^-, d)$ as 
$$P(h_1 | h^-, d) = \frac{P(h_1 | h^-, d)}{P(h_1 | h^-, d) + P(h_2 | h^-, d)} \propto P(h_1|h^-)$$
$$P(h_2 | h^-, d) = \frac{P(h_1 | h^-, d)}{P(h_1 | h^-, d) + P(h_2 | h^-, d)} \propto P(h_2|h^-)$$

\noindent 
where we define $h^-$ to be the set of all words that are shared between $h_1$ and $h_2$; note that this is the union of the words in $\beta$ and $\delta$. Then we explicitly compute each of the above conditional probabilities as follows. Note that in the interest of space, we omit most of the derivations, which can be found in the paper by Goldwater \textit{et al.}

\begin{align*}
    P(h_1 | h^-) &= P(w_1 |h^-)P(u_{w_1} | h^-) \\
                &= \frac{n_{w_1}^{(h^-)} + \alpha_0 P_0(w_1)}{n^- + \alpha_0} \cdot \frac{n_{u}^{(h^-)} + \frac{\rho}{2}}{n^- + \rho} \\
    P(h_2 | h^-) &= P(w_2, w_3 | h^-) \\
                &= P(w_2 | h^-)P(u_{w_2} | h^-)\\
                &*P(w_3 | w_2, h^-)P(u_{w_3} | u_{w_2}, h^-) \\
                &= \frac{n_{w_2}^{(h^-)} + \alpha_0 P_0(w_2)}{n^- + \alpha_0} \cdot \frac{n^- - n_{\$}^{(h^-)} + \frac{\rho}{2}}{n^- + \rho} \\
                & \cdot \frac{n_{w_3}^{(h^-)} + I(w_2 = w_3) +\alpha_0 P_0(w_3)}{n^- + 1 + \alpha_0} \\
                & \cdot \frac{n_u^{h^-} + I(u_{w_2} = u_{w_3}) + \frac{\rho}{2}}{n^- + 1 + \rho}
\end{align*}

\noindent 
We define $n^-$ as the total number of words in $h^-$ and $n_{w_1}^{(h^-)}$ as the number of occurrences of $w_1$ in $h^-$; $n_{w_2}^{(h^-)}$ and $n_{w_3}^{(h^-)}$ are defined similarly. Furthermore, $n_{\$}^{h^-}$ is the total number of utterance boundaries (and thus utterances) in $h^-$, and $n_u^{(h^-)} = n_{\$}$ if $w_1$ is utterance-final and $n_u^{(h^-)} = n^- - n_{\$}$ otherwise. $I(\cdot)$ is an indicator function that equals $1$ when its argument is true and 0 otherwise. $P_0$ corresponds to the distribution described in step 2a of unigram generative model.   

An example of how the Gibbs sampling algorithm works in the unigram model is shown in Figure \ref{fig:sample-gibbs}. This image is taken from the Goldwater paper on which we have based our research \cite{goldwater}.

\begin{figure*}[t]
    \centering
    \includegraphics[width=0.7\textwidth]{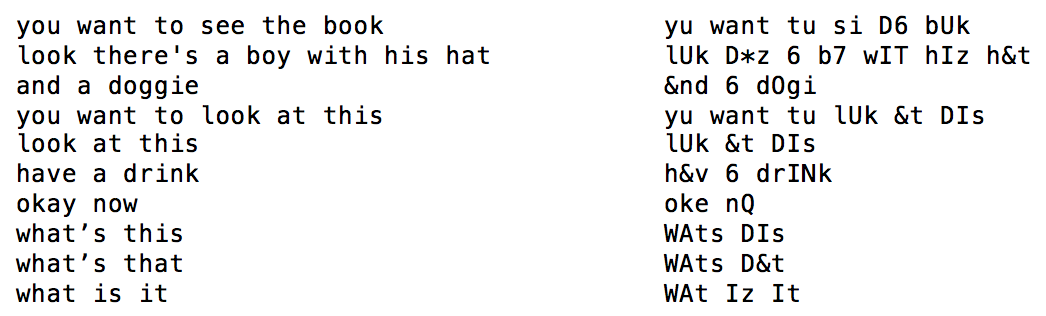}
    \caption{Sample utterances and their phonetic representation.}
    \label{fig:sample-utterances}
\end{figure*}

\subsection{Inference Under Bigram Model}
Similar to the unigram model, we have two possible segmentations for the bigram model: $s_1=\beta w_lw_1w_r\gamma$ and $s_2=\beta w_lw_2w_3w_r\gamma$, where $w_l$ and $w_r$ are the left and right context words respectively, and $\beta$ and $\gamma$ are the remaining words. Then, we can compute $P(s=s_1|h^-)$ and $P(s=s_2|h^-)$ as
\begin{align*}
P(s=s_1|h^-) &= \frac{n^{h^-}_{\langle w_1,w_1\rangle}+\alpha_1P_1(w_1|h^-)}{n^{h^-}_{w_l}+\alpha_1}\\
&\cdot \frac{n^{h^-}_{\langle w_1,w_r\rangle}+\alpha_1P_1(w_r|h^-)}{n^{h^-}_{w_l}+\alpha_1} \\
P(s=s_2|h^-) &= \frac{n^{h^-}_{\langle w_1,w_2\rangle}+\alpha_1P_1(w_2|h^-)}{n^{h^-}_{w_l}+\alpha_1}\\
&\cdot \frac{n^{h^-}_{\langle w_2,w_3\rangle}+\alpha_1P_1(w_3|h^-)}{n^{h^-}_{w_2}+\alpha_1} \\
& \cdot \frac{n^{h^-}_{\langle w_3,w_r\rangle}+\alpha_1P_1(w_r|h^-)}{n^{h^-}_{w_3}+\alpha_1} 
\end{align*}
The derivations have been omitted from this paper, as they are discussed in detail by Goldwater \textit{et al.} (Goldwater \textit{et al.}, 2009).

\section{Materials \& Methods}

As introduced previously, our simulations and data are based off of the paper by Goldwater \textit{et al.}

\subsection{Data}
We obtained a corpus of 9790 lines of transcribed child-directed speech derived from the  Bernstein-Ratner corpus in the CHILDES database; these utterances are specifically directed towards 13- to 23-month-olds \cite{br, brentdata, childes}. This is the same corpus used by Goldwater \textit{et al.}, in addition to other notable papers in the same field such as ones by Brent (1999) and Venkataraman (2001) \cite{brent, ngs}. The corpus includes the original text of the child-directed speech and the equivalent phonetic representation. The first ten lines of the original text (post-processed by Brent) along with the phonetic transcription is given in Figure \ref{fig:sample-utterances}. For our inference models, we use the phonetic version of the child-directed speech as input. On average, each utterance contains 3.41 words and each word contains 2.87 phonemes. In our model, we provide the utterance boundaries and attempt to learn all the word boundaries. 

\subsection{Implementation \& Code}
We implemented a Gibbs sampling procedure in Python for both the unigram and bigram models, based off the descriptions provided in Section 4. The input was the first 100 lines of the Bernstein-Ratner corpus. (See Section 5.4 for further discussion on modifications we made to the experiment procedures used by Goldwater \textit{et al.}) We ran the sampling procedure for 11,000 iterations, 1000 for the ``burn-in'' period and another 10,000 where we collected the actual samples. In total, we collected 1000 samples, one every ten iterations. This is because our Gibbs sampling procedure only makes local changes during each pass through the corpus, so we assume that the change in word boundary distributions will be minimal between subsequent iterations. This implies that we can afford to sample from the current distribution less frequently. 

To achieve faster convergence, we incorporated the concept of annealing to our sampling procedure \cite{annealing}. For this process, we introduce a temperature parameter $\gamma$ that starts high and is gradually reduced to 1, and we raise all our computed probabilities $P(h|d)$ to $\frac{1}{\gamma}$. This helps make the sampled distribution more uniform when $\gamma > 1$, and thus expedites the process of exploring a larger sample space \cite{annealing}. As a result, we are more likely to transition to the true distribution more quickly.

\subsection{Evaluation Metrics} 
We employ the same metrics used by Goldwater \textit{et al.} to conduct an evaluation of our model. Specifically, we look at precision (the number of true positives over the total number of discovered positives) and recall (the fraction of discovered positives over all positives) \cite{goldwater}. In addition, we also consider the $F_0$ score, which is the geometric average of precision and recall, e.g. we define 
$$F_0 = \frac{2 \cdot \text{precision} \cdot \text{recall}}{ \text{precision} + \text{recall}}$$

\noindent 
In our evaluation, we considered the precision ($P$), recall ($R$), and $F_0$ score (F) on words in the context of the corpus. In other words, we required both boundaries for a word to be correct in order for the word to be classified as correct. We also considered the precision ($LP$), recall ($LR$), and $F_0$ ($LF$) on the lexicon, e.g. the set of unique word types discovered by the sampling procedure. Finally, we evaluate the location accuracy of each word boundary by computing the corresponding precision ($BP$), recall ($BR$), and $F_0$ scores ($BF$) \cite{goldwater}.

\subsection{Challenges \& Modifications}

While researching and implementing the Gibbs sampling procedure presented by Goldwater \textit{et al.}, we ran into a number of obstacles that we attempted to solve by introducing modifications to the original model. These challenges and subsequent modifications are discussed below. 

\subsubsection{Slow Convergence}
The Gibbs sampling procedure that we implemented has the disadvantage of being slow to converge, since the probability that the input is randomly initialized to a segmentation in or near the solution space is low. Furthermore, each iteration through the input only looks locally at one boundary at a time, so transitioning from one hypothesis to another is difficult. In particular, these transitions may require movement through low-probability intermediate hypotheses.

One approach we used to achieve faster convergence was to introduce the temperature parameter $\gamma$ as described previously in section 5.2. Even so, we were unable to match the results achieved by Goldwater \textit{et al.}. We hypothesize that this is due to the fact that we did not run the sampling procedure for enough iterations. We tested this idea by initializing all word boundaries correctly except for a few and examined the segmentation that was produced after performing inference on this input. Holding the number of iterations constant, the accuracy of the segmentation decreased as we increased the number of incorrectly placed boundaries to an extent before plateauing. (See Section 6.4 for details.)

\begin{table*}[t]
    \begin{center}
        \begin{tabular}{|c | c c c | c c  c | c c c |}
        \hline 
        Model & P & R & F & BP & BR & BF & LP & LR & LF \\
        \hline 
        DP & 0.619 & 0.476 & 0.538 & 0.924 & 0.622 & 0.743 & 0.570 & 0.575 & 0.572 \\
        DP' & 0.682 & 0.543 & 0.605 & 0.696 & 0.482 & 0.570 & 0.472 & 0.644 & 0.545 \\
        \hline 
        \end{tabular}
    \end{center}
    \caption{Comparison of the results from our DP model to the results of Goldwater \textit{et al.}}
    \label{tab:unigram-results}
\end{table*}

\begin{figure*}[b]
    \centering
    \includegraphics[width=0.7\textwidth]{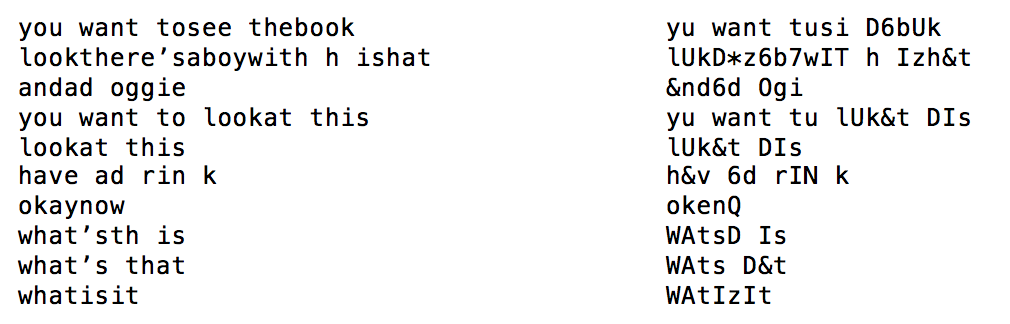}
    \caption{First ten lines of the corpus segmented using our inference model.}
    \label{fig:sample-inference}
\end{figure*}

\subsubsection{Large Input}
Recall that the input is derived from 9790 utterances, with  3.41 words per utterance and 2.87 phonemes per word on average. This means that the number of potential boundary locations we consider on every iteration close to 100,000. Goldwater \textit{et al.} ran their sampling procedure for 20,000 iterations, in addition to a couple thousand additional iterations comprising the ``burn-in'' period \cite{goldwater}. This amounted to an unwieldy experiment that we did not have the computational resources to carry out. Thus, in our project, we only ran the Gibbs sampling procedure for 11,000 iterations, 1000 for the ``burn-in'' period and another 10,000 where we collected the actual samples (see Section 5.2).

In addition, we only used the first 100 lines in the corpus to perform inference. Though the time to process 100 lines was still relatively lengthy, it was much more manageable than using the entire input corpus. The disadvantage to using only a subset of the input is that the word type and phoneme type distribution may not match that of the true corpus, but we are not too concerned about this because inference on the partial input yielded fairly reasonable results. In addition, when analyzing different parameter settings and the impact of varying the number of incorrect boundaries, we used a toy corpus consisting of the first 10 lines of the original data set. 

\subsection{Oversegmentation}
The model by Goldwater \textit{et. al.} assumed a uniform distribution across phonemes \cite{goldwater}. In other words, when computing $P(w_i = x_1,...,x_M|w_i) = p_{\#}(1-p_{\#}^{M-1}\Pi_{j-1}^MP(x_j)$ in step 2a of the generative process, the value for $P(x_j)$ for every possible phoneme $x_j$ was the same. We noticed that the assumption that each phoneme is uttered with the same probability caused the corpus to be oversegmented, since this setup suggests that shorter words are favored. 

Both the MBDP-1 and NGS models incorporate the idea that phonemes have different probabilities of occurrence \cite{brent, ngs}. The NGS model estimates phoneme probabilities according to their empirical distribution in the corpus \cite{ngs}, and this is precisely the appproach we use in our implementation of Goldwater's DP model. 

Specifically, we use a distribution where $P(x_j) = n_{x_j}/n,$ where $n_{x_j}$ is the number of times the phoneme $x_j$ occurs in the corpus and $n$ is the total number of phonemes in the corpus. Thus, in our models, a more common phoneme is more likely to be generated in the word $w_i$ than a less common phoneme.

\section{Results \& Discussion}
Although we were unable to match most of the numbers reported by Goldwater \textit{et al.} in their paper, the general trends in our results seem to be consistent with theirs. However, we also produced some results that were different, which we analyze below. Most of this section focuses on the unigram model.

\subsection{Segmented Corpus}
We report the metrics for the corpus segmented by our Gibbs sampling procedure under the unigram model in Table \ref{tab:unigram-results}, along with the results obtained by Goldwater \textit{et al.} We refer to their model as the DP model and ours as DP'. Since the analysis in Goldwater \textit{et al.} focuses on results where $p_{\#} = 0.5$ and $\alpha_0 = 20$, we do the same here \cite{goldwater}. A discussion on the effects of varying these parameters can be found in Section 6.2.

\begin{figure*}[t]
    \centering
    \includegraphics[width=0.7\textwidth]{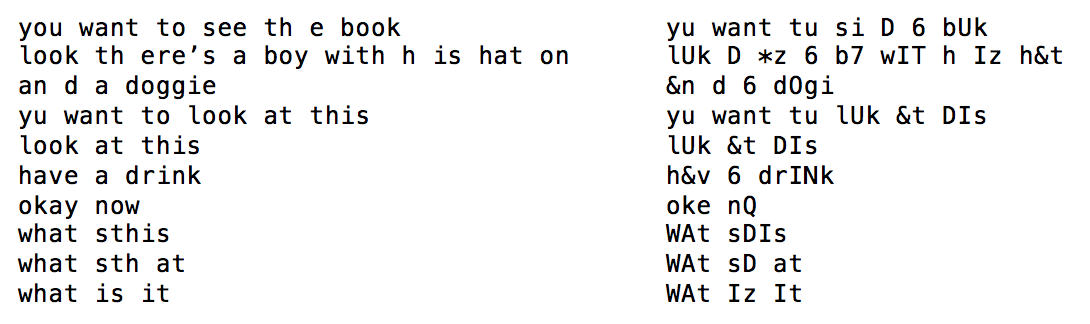}
    \caption{Result of sampling using only the first ten lines of the corpus.}
    \label{fig:train-ten}
\end{figure*} 

Looking at Table \ref{tab:unigram-results}, we see that our overall precision, accuracy, and recall (on words) are all slightly higher than Goldwater \textit{et al.}, but our performance on potentially ambiguous boundaries was substantially worse. Indeed, words that frequently co-occur together are often undersegmented as one word. In Figure \ref{fig:sample-inference}, which contains the segmentation produced by our algorithm for the first 10 utterances in the data set, we see that phrases like ``thebook'', ``lookat'' and ``tosee'' are regarded as one word. This makes sense because the two words comprising each of these phrases generally occur together. Given that our model takes into consideration the number of occurrences of a given word, if a certain phrase occurs multiple times, then it is more likely that the model will think that that phrase is a single word.

In addition to undersegmentation of collocated words, we also see some instances of oversegmentation. In Figure \ref{fig:top-20}, we provide the 20 most frequently occurring words in the first 100 utterances of the corpus, in addition to the top 20 words that we discovered in the input text. Note that there are many one-phoneme words among the top words we discovered, most of which are not actual words on their own. (The exception is ``6'' and ``9'', which represent ``a'' and ``I'', respectively, when they are not joined with other phonemes.) This suggests that these phonemes may occur frequently in words that are common in the corpus. 

For example, consider the the phoneme ``D''. This phoneme often occurs at the beginning of a word like ``this'' (``DIs'') or ``that'' (``D\&t''), where if we remove the ``D'', the remaining word is still a valid word. In this case, we would get ``is'' (``Is'') and ``at'' (``\&t''), respectively. Since $p_\# = 0.5$, and thus $1 - p_\#$ is also equal to $0.5$, it is not unreasonable to assume that short words are favored by our model, which may lead to a segmentation where a phoneme like ``D'' is split from the true word. Even so, we note that our model overall tends to undersegment words, which can be seen by the long words in Figure \ref{fig:sample-inference}. Furthermore, we computed the average word length in phonemes for the first 100 utterances in the corpus and compared it to the average word length in phonemes for the words discovered by our sampling procedure. The actual average word length was about 2.77 but our discovered length was about 3.48, leading us to conclude that our model undersegments words.

\begin{center}
    \includegraphics[width=0.27\textwidth]{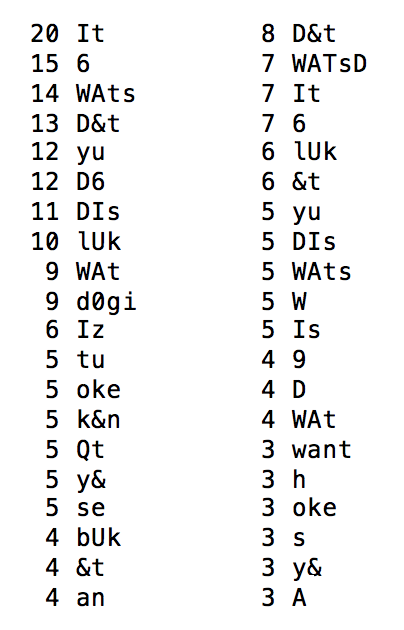}
    \captionof{figure}{Comparison of the most frequent words discovered by our model to the top-occurring words in the first 100 lines of the actual corpus.}
    \label{fig:top-20}
\end{center}

\subsection{Parameter Variation}
Our model had two variables, $p_{\#}$ and $\alpha_0$, that could be varied. We experimented with various values of $p_{\#}$ while holding $\alpha_0 = 20$, and then various values of $\alpha_0$ while keeping $p_{\#}$ at 0.5. The results for each experiment are plotted in Figure \ref{fig:param-variation}. We examine the change in F, BF, and LF produced by changing the values for these two parameters. 

Based on the graphs, we see that there is an apparent optimal value for $\alpha_0$ that occurs around $\alpha_0 = 20$. Values that are too low or too high generally yield less favorable results, although very small values of $\alpha_0$ seem to improve the model performance. Note that the value of $\alpha_0$ influences the number of new words that are discovered in the model; as a result, we hypothesize that larger values of $\alpha_0$ decrease the $F_0$ score due to a decrease in precision. This is reasonable because the original corpus only contains a limited vocabulary; a high capacity to discover new words is not advantageous to the model's performance. 

\begin{center}
    \includegraphics[width=0.48\textwidth]{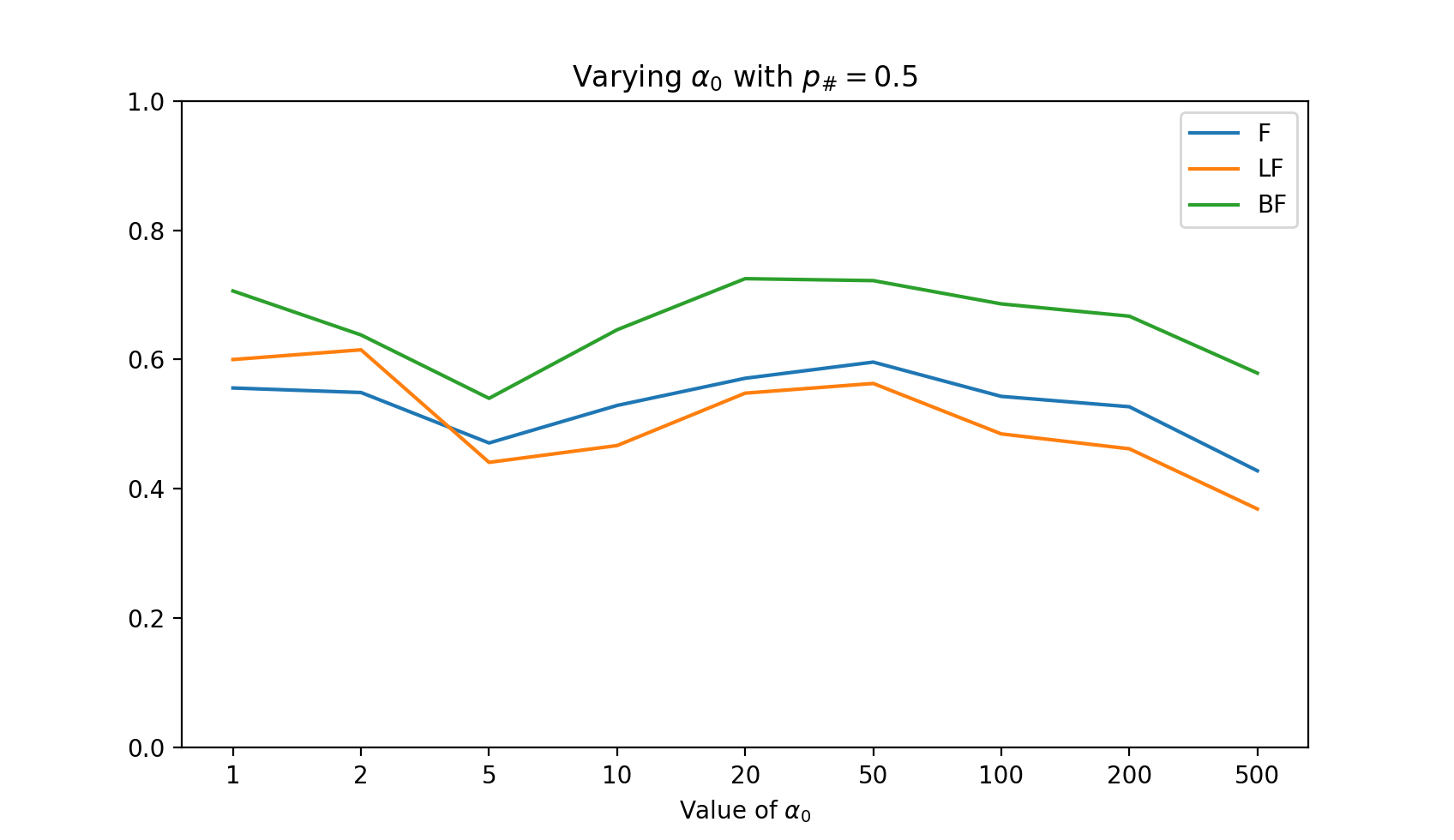}
    \includegraphics[width=0.48\textwidth]{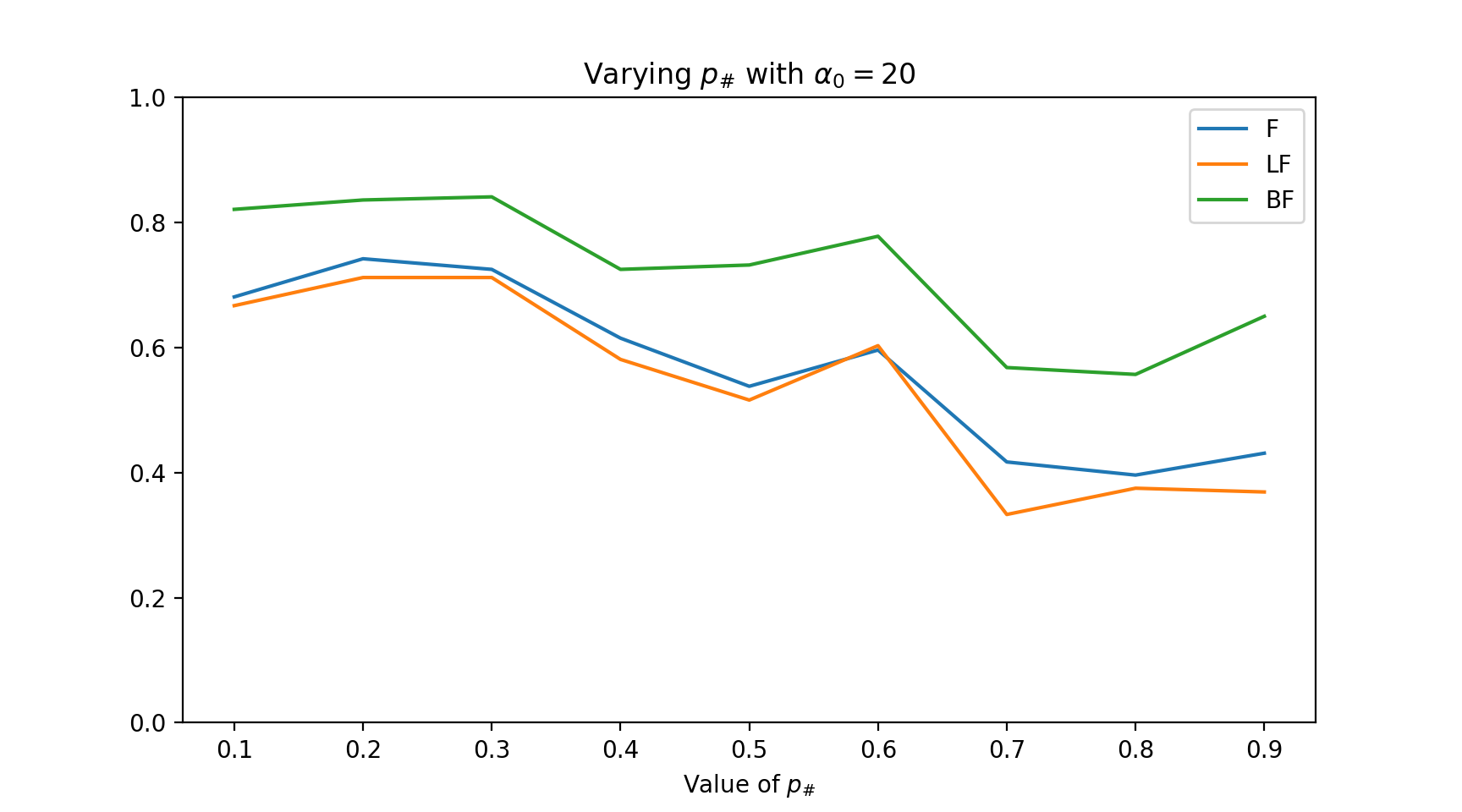}    \captionof{figure}{\textit{Top.} The effect of varying $\alpha_0$ on model performance. \textit{Bottom.} The effect of varying $p_{\#}$ on model performance. We specifically look at the $F_0$ score on words, lexicon, and boundaries.}
    \label{fig:param-variation}
\end{center}

\noindent 
For $p_{\#}$, we find that the overall trend is a decrease in F, LF, and LR for increasing values of $p_{\#}$. Lower values of $p_{\#}$ favor longer words, which thus improves the $F_0$ score by improving recall. This means that we as we increase $p_{\#}$, recall degrades and we produce shorter words.  

\subsection{Toy Corpus}
Recall that  we run the Gibbs sampling procedure for less than half the number of iterations compared to the experiment by Goldwater \textit{et al.} As a result, we hypothesize that one reason for our model's relatively worse performance is that we didn't allow for enough time to converge. To test this idea, we ran the same sampling procedure on the first 10 lines of the input corpus only and display the resulting segmentation in Figure \ref{fig:train-ten}.  

We can see that the segmentation is more accurate than the one presented in Figure 3 that used 100 utterances. There is much less undersegmentation than previously, and in fact, there appears to be more oversegmentation. However, half of the utterances are now segmented exactly correctly, which supports the hypothesis that our sampling algorithm is eventually able to reach convergence, but that that convergence may take a very long time to achieve. 

\subsection{``Nice'' Initialization}
Another way in which we tested the correctness of our sampling procedure was by providing a ``nice'' initialization. In other words, we set all the word boundaries correctly and then iterate through the text $k$ times, each time flipping the boundary at a random index. In other words, if there is a current boundary at some index $i$, we get rid of it; if there is no boundary, then we introduce a boundary. Note that with this procedure, it is possible to select the same index multiple times and thus switch the boundary multiple times. However, given the number of indices under consideration, we assume that this is not a problem until $k$ grows large. Once we finish the boundary-flipping procedure, we use the resulting text as the input to our sampling procedure. We tested various values of $k$ and plot the results in Figure \ref{fig:incorrect-boundaries}.

\begin{center}
    \includegraphics[width=0.48\textwidth]{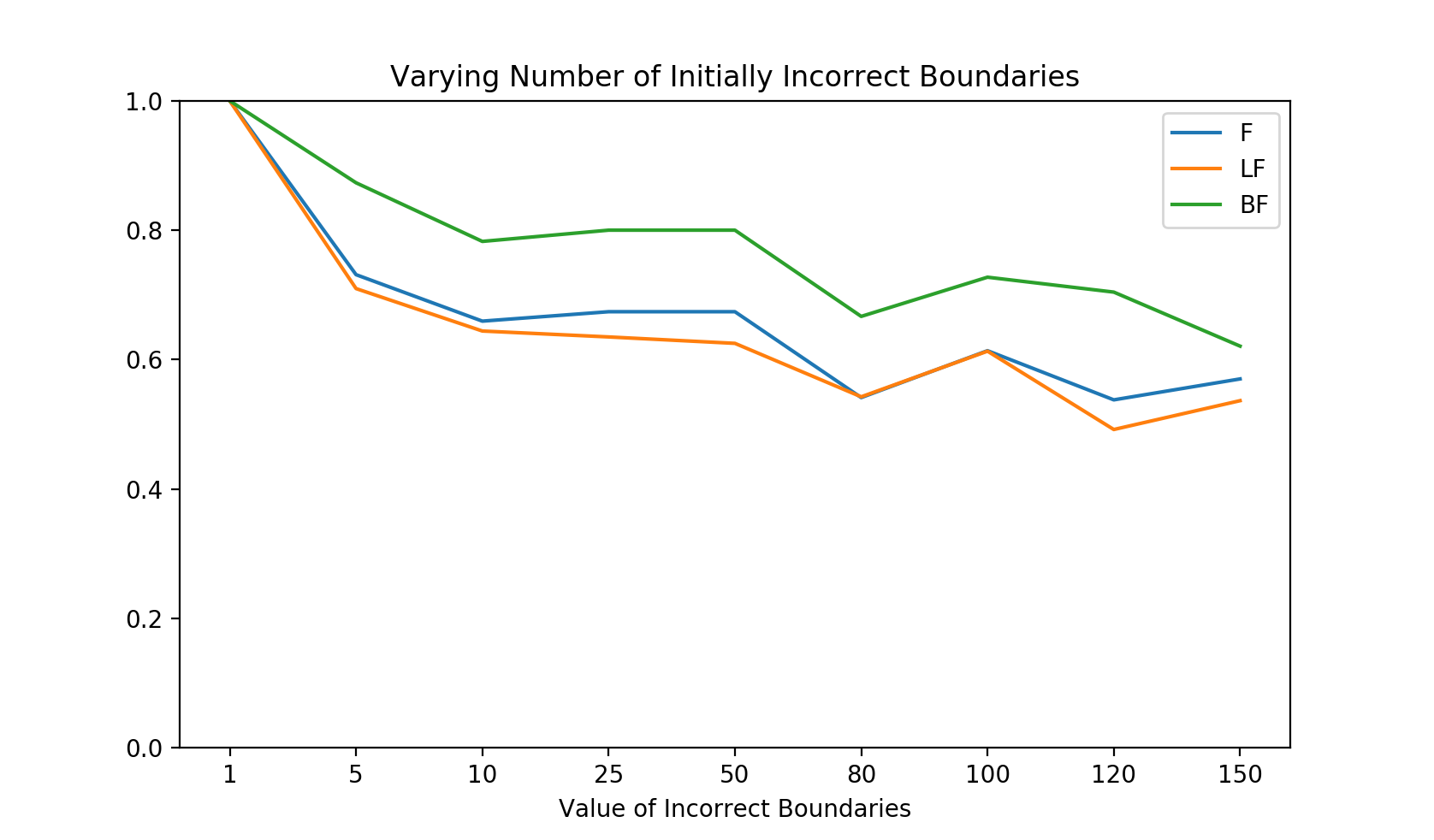}  
    \captionof{figure}{The effect of increasing the number of incorrectly initialized boundaries on model performance.}
    \label{fig:incorrect-boundaries}
\end{center}

We see that with only one incorrectly initialized boundary, we are able to segment the corpus exactly; however, as we increase the number of incorrectly placed boundaries, the performance of the model decreases. When $k$ grows close to or larger than the total number of phonemes in the input, the values of F, LF, and BF seem to flatten out and approach some constant. This makes sense because with a large $k$, we are essentially performing a random initialization, which is what we used in our original model. 
\subsection{Prior Word Knowledge}
Studies show that by the end of 12 months, infants generally know how to say a few basic words and understand what they are saying \cite{goldwater}. Since the data that we are using in our model is speech directed towards children between 13 and 23 months, we thought it would make sense to introduce a vocabulary to the model and examine the results. This simulates the idea that children can use previously learned words to figure out how to segment a new utterance. 

To implement this extension, we added an additional parameter to the model, the size of the learned vocabulary $v$, and a ``boost'' factor $b$. We tested various values of $v$, where a small value corresponds to a (typically younger) child who only knows a few words and a large value corresponds to a (typically older) child who has learned a larger vocabulary. Prior to sampling, we randomly select $v$ words from the set of words in the input with probability proportional to their frequency in the set of input utterances. Then, when considering a potential boundary location at index $i$, if $w_1$ is a word in the vocabulary, we amplify the probability $P(h_1 | d)$ by some factor $b$ so that it is more likely to be chosen. Similarly, if $w_2$ and $w_3$ are words in the vocabulary, we amplify $P(h_2 |d)$ by $b$. (Recall that $w_1$, $w_2$, and $w_3$ are defined in Section 4.2.) We multiply a hypothesis by $b$ instead of explicitly choosing that hypothesis in order to model the idea that children may incorrectly segment words that sound similar or words that are merged with other words, for example as contractions or compound words. Setting $b$ to infinity is equivalent to the setting where a child knows for certain how to correctly segment all the words in their learned vocabulary. The results are shown in Figure \ref{fig:partial-vocab}.

\begin{center}
    \includegraphics[width=\linewidth]{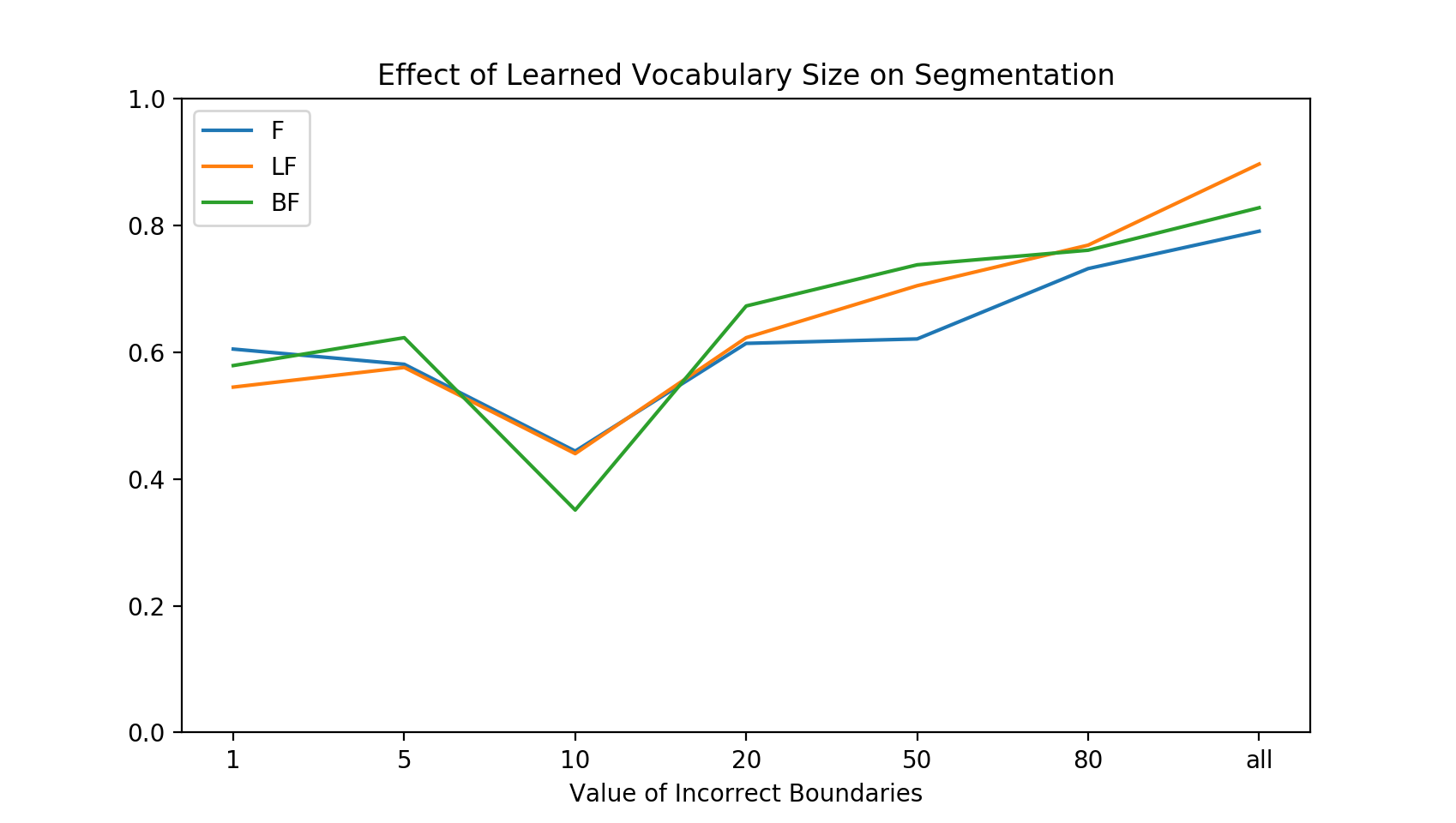}  
    \captionof{figure}{Results of adding a partial learned vocabulary to the model.}
    \label{fig:partial-vocab}
\end{center}

\begin{figure*}[b]
    \centering
    \includegraphics[width=0.7\textwidth]{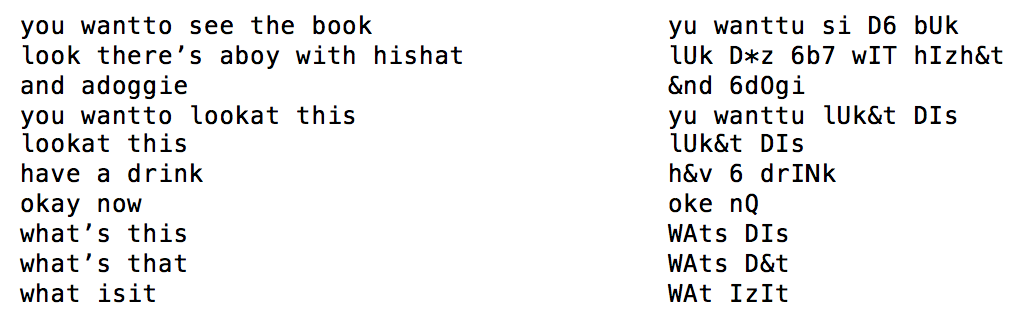}
    \caption{First ten lines of the corpus segmented using our bigram inference model.}
    \label{fig:bigram-segmentation}
\end{figure*}

\noindent 
We see that in general, as the size of the vocabulary grows, the performance of the model improves substantially. In particular, the $F_0$ score on the lexicon increases the most, which makes sense given that introducing a partial learned vocabulary encourages the model to recognize words that exist in that vocabulary. However, we notice the sudden drop in performance with a vocabulary size 10. We hypothesize that some words that can occur in contractions or as part of compound words may be oversegmented in order to extract a word that exists in the vocabulary. For example, if ``what'' is in the vocabulary, our model may try to segment all occurrences of ``what's'' as ``what'' and ``s''. Similarly, if ``it'' or ``is'' is in the vocabulary, all words that contain these two words are also more likely to be segmented incorrectly. But as the size of the vocabulary grows, the accuracy of our model increases because the likelihood that we see contractions, compound words, etc. is larger. In other words, for example, it is more likely that both ``what'' and ``what's'' occur in the vocabulary, so the model doesn't accidentally oversegment ``what's''.

However, note that even when the vocabulary contains all the words in the lexicon, we still do not get 100\% accuracy. We propose that this may be because we did not let our sampling algorithm run long enough or that in some cases, it may be more ``ideal'' (e.g. higher probability) for the model to segment a word incorrectly in order to produce another word that is in the vocabulary. 

Finally, we recognize that randomly initializing a size $v$ vocabulary from the words in the input data may not accurately reflect a young child's true knowledge, since some words commonly learned during the first two years of life are more typical than others. Even though we choose words based on their relative frequency in the corpus, we point out that the corpus contains words uttered by the parents rather than the children; however, it is not unreasonable to assume that a child might pick up on some of these words if the hear their parents say them frequently. In any case, it would be interesting to experiment further with the idea of a partial vocabulary, perhaps incorporating a more sophisticated model.

\subsection{Bigram Model}
We ran our bigram model implementation using the same number of iterations as the unigram model; however, we introduce an additional parameter $\alpha_1$ as described in Section 4.3. For our experiments, we use $\alpha_1 = 100$, $\alpha_0 = 3000$, and $p_{\#} = 0.2$, which are the same settings as in Goldwater \textit{et al.} A sample of the segmentation is provided in Figure \ref{fig:bigram-segmentation}. 

Qualitatively, we see that the segmentation is improved compare to the unigram model. Specifically, we no longer have words that are oversegmented, e.g. single words that are split into multiple words by the model. However, we still see undersegmentation, where two words are combined into one, especially for multi-word phases. For example, the phrases ``want to'' and ``look at'' are frequently interpreted as single words. But we also note that ``want to'' often follows ``you'' while ``look at'' occurs right before ``this''. This suggests that the bigram model is interpreting ``you want to'' and ``look at this'' as the bigrams ``you wantto'' and ``lookat this''. If these phrases occur often enough in the corpus, then it is not unreasonable for this kind of undersegmentation to result. In Section 7, we will discuss using a trigram to better segment such phrases.

\section{Future Work}
Our results and that of Goldwater \textit{et al.} indicate that context helps in learning word segmentation. The unigram model tends to undersegment the corpus, especially two or three word phrases (Goldwater \textit{et al.}, 2009). While the bigram model is better at segmenting the corpus, we hypothesize that a trigram model will outperform both the unigram and bigram model. In particular, it should be better at segmenting common three word phrases than the unigram and bigram models. However, limits on computational resources currently prevent us from testing such a trigram model.

Furthermore, we propose an alternative model to using context in learning word boundaries: incorporating the use of posteriors to reevaluate word boundaries. That is, given the word $w_{i+1}$, identify boundaries in the word $w_{i}$ similar to the process in the bigram model. The intuition is that for undersegmented phrases such as "ofthe cat", this model can use the context "cat" to more easily identify the boundary between "of" and "the." We hypothesize that a bidirectional bigram model will produce better results than our current model. However, this model will be computationally expensive and will require greater computational resources than we currently have. 

Finally, we acknowledge that our models do not perfectly simulate infants as learners. Unlike infants, our models are essentially perfect learners: they will produce ideal results given an input. Furthermore, infants sometimes hear utterances that share common words (e.g. common sequences of phonemes), even though they may not understand the meaning of the word itself. Thus, it would be interesting to develop a model that incorporates some prior knowledge on previous utterances and examine how it might perform on segmentation of new utterances.

\section{Conclusion}
To better learn how children learn word boundaries, we implemented two types of models based on the work of Goldwater \textit{et al.} to test the importance of context in learning word segmentation. Our unigram model tests the assumption that words are generated independently and have no predictive power. Our bigram model tests the assumption that words can be used to predict other words. Although our models did not generate the same numbers reported by Goldwater \textit{et al.}, our models exhibited the same general trend: the bigram model is better at learning word boundaries than the unigram model. From this we concluded that context does appear to help children learn word segmentation. For further research, we propose to investigate the performance of a trigram model, a bidirectional bigram model, as well as a model with a more realistic learner.

\end{multicols*}
\end{document}